\documentclass{article}
\usepackage[preprint]{neurips_2026}
\setcitestyle{numbers,square,comma}
\usepackage[T1]{fontenc}
\usepackage[utf8]{inputenc}
\usepackage{graphicx}
\usepackage{booktabs}
\usepackage{amsmath}
\usepackage{hyperref}
\hypersetup{colorlinks=true,linkcolor=blue,citecolor=blue,urlcolor=blue}
\makeatletter
\renewcommand{\@notice}{}
\makeatother
\title{How Seasonal Reference Choice Changes the Apparent Gains of Chronos Models in Electricity Load Forecasting}
\author{ChandraMohan K\\ Bannari Amman Institute of Technology, Sathyamangalam, Tamil Nadu, India}
\begin{document}
\maketitle

\section*{Abstract}

When a forecasting model is said to beat a simple baseline, the size of that win depends on which baseline was used. We measure how much this matters for two zero-shot foundation models, Chronos-Bolt and Chronos-2, on hourly electricity load. The study uses 60 fixed cases drawn from eight series in four public collections, with forecasts 24 and 168 hours ahead and up to 2,048 hours of history. Each model is compared with daily persistence, weekly persistence, a weekday hybrid, a daily-weekly blend, and a reference chosen from these on a validation period. We keep every origin-level forecast and estimate uncertainty with paired circular block bootstrap intervals using 168- and 336-hour blocks. A week ahead, Bolt clearly beats weekly persistence in 21 of 30 cases but the validation-selected reference in only 14, and its median gain falls from 17.3\% to 11.9\%. For Chronos-2 the clear-win counts are 18 and 15. Chronos-2 has lower week-ahead MAE than Bolt in 24 cases, yet only 12 of those differences are clear. The selected reference is not always the harder baseline, since on some test windows it does worse than weekly persistence. The choice of baseline therefore changes both how large the reported gain is and how sure we can be of it. The results are exploratory: the series and windows are related, an earlier pilot was inspected, pretraining overlap cannot be ruled out, and the case-level intervals are not corrected for multiple comparisons.

\noindent\textbf{Keywords:}  electricity load forecasting; time-series foundation models; Chronos; seasonal persistence; benchmark design; paired bootstrap

\section{Introduction}

A reported improvement in forecast accuracy is always measured against a reference forecast, so its size depends on which reference is used. For hourly electricity load, two simple references are natural: repeat yesterday, or repeat the same hours from last week. Which of the two is harder to beat varies across meters, seasons and horizons. A model compared only with last week can look strong in places where a daily or blended reference would have been close behind it.

This paper looks at how that choice shapes claims about Chronos-Bolt and Chronos-2. We hold the forecasts fixed, including the origins, targets, horizons and the models' own predictions, and change only the seasonal reference. Every reference uses data available at the forecast origin, and the final rule and blend weight are chosen on an earlier validation period before the test period is scored.

Foundation models have already been evaluated on electricity load \cite{ref1,ref2,ref3,ref4}. Simeone tests Chronos-Bolt and Chronos-2 with long contexts against a weekly reference \cite{ref1}. Meyer and colleagues benchmark several models on household load \cite{ref2}, Vasulkar and colleagues compare daily and weekly references in a single building \cite{ref3}, and FETS covers a much wider range of energy datasets \cite{ref4}. Our aim is narrower: we measure how a reference chosen without looking at the test data changes the reported gain and its paired uncertainty.

We ask three questions. How do the two models compare with weekly and validation-selected references one and seven days ahead? How often does switching the reference turn a clear win into an unclear one? And how do Chronos-2 and Bolt compare when they see exactly the same origins and histories?

\section{Related Work}

The closest study is Simeone \cite{ref1}, which evaluates several foundation models, Bolt and Chronos-2 among them, on ERCOT hourly load with 24- and 168-hour horizons and contexts of up to 2,048 hours. Our model and horizon setup is very similar, but Simeone uses weekly persistence throughout, while we vary the reference on the same held-out predictions and pick one rule on validation data.

Meyer et al. \cite{ref2} study several household load datasets with a range of foundation and trained models. Their seasonal reference averages past days instead of copying the last observed day. Vasulkar et al. \cite{ref3} are the closest in motivation. They report both daily and weekly references for one building and compare models with paired uncertainty, but only for day-ahead forecasting, and without a reference chosen on validation data or a 168-hour horizon across several load collections. FETS \cite{ref4} has already made broad foundation-model benchmarking in energy an established topic, and we do not claim that as new.

Reference choice has also been studied outside foundation-model work. In solar forecasting, Yang \cite{ref5} derives the optimal convex combination of climatology and persistence and shows that the best reference depends on the horizon. In day-ahead electricity price forecasting, Lago et al. \cite{ref6} list three natural naive references: the previous day, the previous week, and a hybrid that takes the previous week for Saturday, Sunday and Monday and the previous day otherwise. They also point out that switching between these references does not change how a fixed set of models ranks. Our blend and hybrid candidates come from these two papers. What we add is to apply them to load forecasting with foundation models, to choose among them on validation data, and to measure how the choice changes the size and certainty of the reported gain, which a ranking alone does not show.

Given this prior work, we make no broad novelty claim about load forecasting or seasonal baselines. Our findings also depend on our candidate rules, our validation window and the two Chronos models we test.

\section{Data and Frozen Evaluation Protocol}

\subsection{Series, periods, and cases}

We use eight hourly series from four public collections: the three zones of the UCI Tetouan power consumption data \cite{ref7}, meters T1 to T3 from the Monash hourly version of ElectricityLoadDiagrams20112014 \cite{ref8}, Victoria regional demand from the Monash elecdemand collection \cite{ref9}, and the Sceaux household series from the UCI household power consumption data \cite{ref10}. The evaluation year is 2017 for Tetouan, 2014 for the Portugal meters and Victoria, and 2009 for the household. Because units differ between collections, we never pool raw MAE across series.

Each series is evaluated in four windows that end at the four calendar quarters, Q1 to Q4. A window has a 42-day validation period, a 168-hour embargo so that no validation target falls inside the test period, and a 42-day test period. Forecast origins move forward one hour at a time. The household series has too many gaps to give complete origins in Q1 and Q2 under our fixed gap rule, so those quarters have seven series per horizon while Q3 and Q4 have eight. That gives 30 cases per horizon and 60 in total. A complete case has 840 test origins; household gaps shorten two of the later cases, and the total across all cases is 48,006 origins. These forecasts overlap heavily, so they are far from 48,006 independent observations. Q4 also repeats the test period of our original pilot and cannot count as independent confirmation.

The inputs were frozen before the long-context and Chronos-2 runs, but the four-window design itself came after we had looked at the pilot results, so this is an exploratory, retrospective study. Input manifests, validation choices, test origins, targets and seasonal forecasts are saved with hashes, and the Chronos-2 outputs were checked against all 60 frozen inputs before any analysis.

\subsection{Forecast origins and seasonal references}

At origin $t$, each model forecasts the next $H$ hours, with $H$ equal to 24 or 168, using only values observed up to $t$. The candidate references follow the same rule. Daily persistence repeats the last observed 24-hour profile across the whole horizon. Weekly persistence copies the same hour from the last observed week. The weekday hybrid, which is the third naive reference of Lago et al. \cite{ref6}, uses the weekly value when the target falls on Saturday, Sunday or Monday and the daily value otherwise, and its daily part is also built only from values observed at the origin. The blend, a convex combination in the spirit of Yang \cite{ref5}, is 
\begin{equation}
\hat{y}^{\mathrm{blend}}_{t+h \mid t} = \alpha\,\hat{y}^{\mathrm{daily}}_{t+h \mid t} + (1-\alpha)\,\hat{y}^{\mathrm{weekly}}_{t+h \mid t}, \qquad h = 1, \dots, H,
\end{equation}
 where $\alpha$ is chosen from 0.00 to 1.00 in steps of 0.05 by validation MAE.

For each series, window and horizon we keep the candidate (daily, weekly, hybrid or tuned blend) with the lowest validation MAE and break ties in a fixed, documented order. Neither the rule nor $\alpha$ is changed for the test period. The selection never sees test data, but it can still overfit 42 days of validation data, so we treat it as one reasonable comparator and not as the best possible reference.

\subsection{Foundation-model inference}

Neither model is fine-tuned. We use Chronos-Bolt base \cite{ref11} (commit 5d9f166) and Chronos-2 \cite{ref12} (commit 29ec376), and take the 0.5 quantile as the point forecast. Both models see up to 2,048 hours of history ending at the origin. In Q1 the data start too late for that, so the history there is between 1,153 and 1,992 hours. Older gaps in the household series are passed to the models as missing values, and the most recent 336 hours are always complete. Chronos-2 forecasts each origin as its own univariate task with cross-learning turned off. Neither model receives weather data or any future value. We record package versions, model revisions, run identities and the full matrix of per-origin forecasts.

\subsection{Scores and dependent uncertainty}

MAE is our main accuracy measure. RMSE, WAPE, and MAPE where the targets allow it, are also saved for every case. The gain of a model over reference $B$ is 
\begin{equation}
G_B = 100\left(1 - \frac{\mathrm{MAE}_{\mathrm{model}}}{\mathrm{MAE}_{B}}\right),
\end{equation}
 which is undefined when $\mathrm{MAE}_{B} = 0$. We never average raw errors across datasets with different units, and we report cases with near-zero reference error separately.

For uncertainty, each origin $t$ contributes one number, its mean absolute error over the horizon, and we take the reference loss minus the model loss: 
\begin{equation}
L_t = \frac{1}{H}\sum_{h=1}^{H} \left| y_{t+h} - \hat{y}_{t+h \mid t} \right|, \qquad d_t = L^{B}_t - L^{\mathrm{model}}_t,
\end{equation}
 so a positive $d_t$ favours the model. We build a 95\% percentile interval for the mean of $d_t$ from 5,000 paired circular block resamples of the hourly origin calendar, once with 168-hour blocks and once with 336-hour blocks. Gaps in the calendar stay as gaps. A comparison counts as clear only if both intervals favour the same method and each block length has at least two blocks' worth of valid origins. Otherwise we call it unclear, or limited by data. We call a model gain meaningful when it is clear and the model's MAE is also at least 5\% lower, and we check 2\% and 10\% as alternatives. The resampling treats the model forecasts and the selected rule as fixed, and we do not correct these per-case intervals for the number of comparisons.

\subsection{Diagnostic local trained models}

We also fit Ridge and LightGBM on each series as a check against local training. Both learn a correction to weekly persistence from past load, recent load statistics and calendar features available at the origin. We choose between 28 and 84 days of recent training origins, three Ridge penalties or two LightGBM leaf settings, using validation MAE. The final model is refitted on mature labels before test and held fixed over the test period. Q1 has no eligible labels before the frozen validation window, so it uses an inner time split within the available pre-test history; training labels end before that inner validation period. All models predict the same frozen hourly test origins as Chronos. This protocol was designed after earlier test results had been inspected, so it is a diagnostic comparison rather than independent confirmation.

\section{Results}

\subsection{Changing the seasonal reference}

Table 1 counts the clear wins. One day ahead, switching to the selected reference hardly matters. Bolt has 23 clear wins against weekly persistence and 22 against the selected reference, and Chronos-2 has 23 and 24. A week ahead, the counts fall: Bolt drops from 21 clear wins to 14 and Chronos-2 from 18 to 15. The cases are related to one another, so these counts describe this fixed benchmark and are not independent replications.

\par\medskip
\begin{center}
\small
\setlength{\tabcolsep}{4pt}
\begin{tabular}{@{}llrr@{}}
\toprule
Horizon & Model & Clear vs weekly & Clear vs selected \\
\midrule
24 h & Chronos-Bolt & 23 / 30 & 22 / 30 \\
24 h & Chronos-2 & 23 / 30 & 24 / 30 \\
168 h & Chronos-Bolt & 21 / 30 & 14 / 30 \\
168 h & Chronos-2 & 18 / 30 & 15 / 30 \\
\bottomrule
\end{tabular}
\end{center}
\noindent\textbf{Table 1:} Cases, out of 30 per horizon, in which the paired bootstrap clearly favours the model: both block lengths agree and there are enough valid origins.
\par\medskip

The selected rule does not beat weekly persistence on every test window. At 168 hours its test MAE is lower in 21 cases, higher in five and equal in four. At 24 hours it is lower in 24, higher in two and equal in four. Switching references can therefore make a model look better as well as worse. In the 168-hour bootstrap, eight Bolt cases are clear wins against weekly persistence but not against the selected reference, while one case that was unclear against weekly persistence becomes clear against the selected reference. For Chronos-2, five cases weaken and two strengthen. This is why the net drop in clear wins is smaller than the number of weakened cases.

The median gain shrinks as well. A week ahead, Bolt's median case-level gain is 17.3\% against weekly persistence and 11.9\% against the selected reference. For Chronos-2 it is 18.7\% and 15.3\% (Table 2). These are medians of case-level ratios, leaving out the undefined ratios where a reference has zero error. They say nothing directly about pooled savings in an energy system or about performance in other years.

\par\medskip
\begin{center}
\small
\setlength{\tabcolsep}{4pt}
\begin{tabular}{@{}llrr@{}}
\toprule
Horizon & Model & Median gain vs weekly & Median gain vs selected \\
\midrule
24 h & Chronos-Bolt & 36.1\% & 18.4\% \\
24 h & Chronos-2 & 39.0\% & 23.4\% \\
168 h & Chronos-Bolt & 17.3\% & 11.9\% \\
168 h & Chronos-2 & 18.7\% & 15.3\% \\
\bottomrule
\end{tabular}
\end{center}
\noindent\textbf{Table 2:} Median case-level MAE gain. Ratios with a zero reference error are undefined and left out; references close to zero can still produce extreme individual ratios.
\par\medskip

\par\medskip\begin{center}
\includegraphics[width=\linewidth]{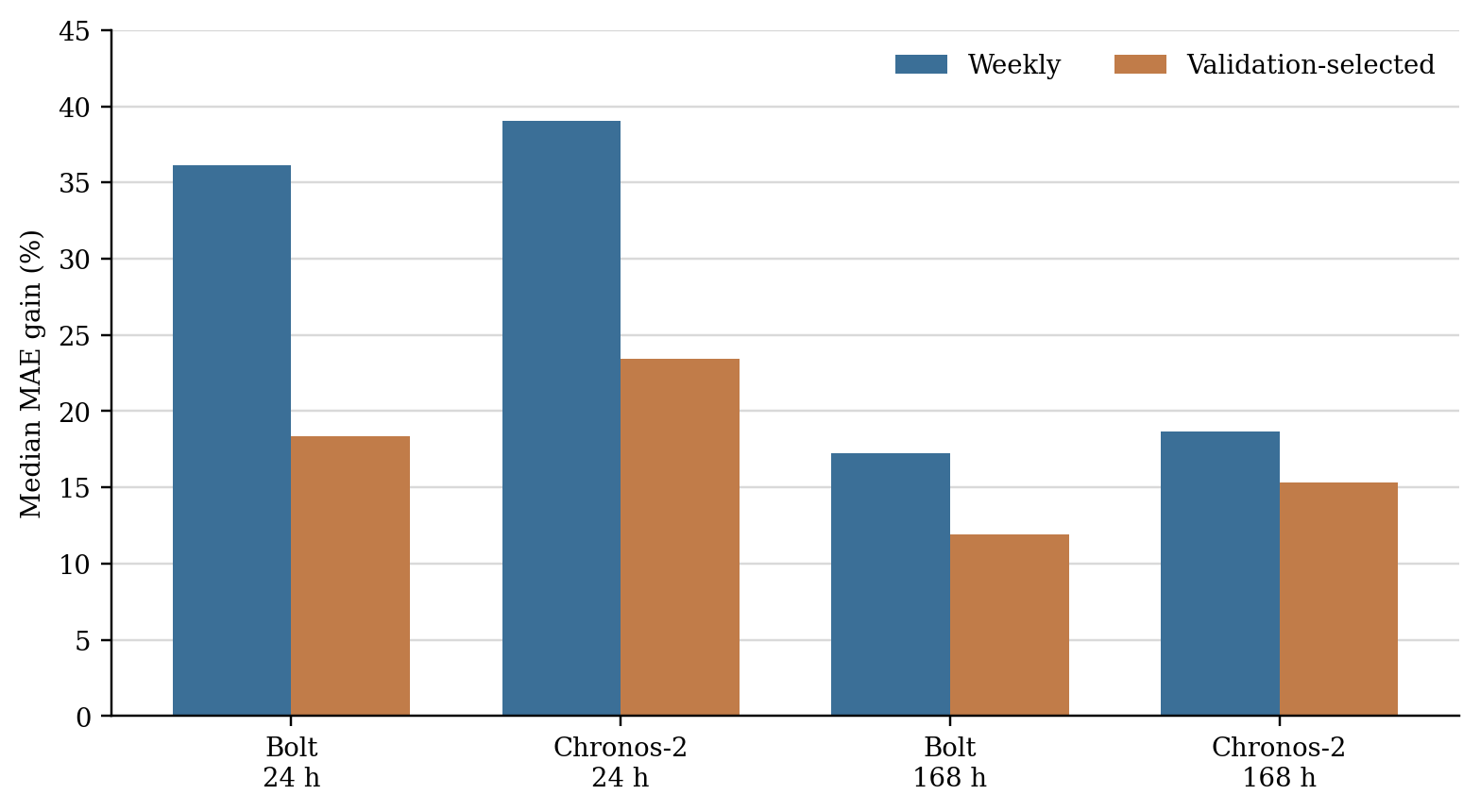}
\end{center}
\noindent\textbf{Figure 1:} Median case-level MAE gain against fixed weekly persistence and against the validation-selected reference. The gain is smaller against the selected reference for both models at both horizons. Load units are not pooled.
\par\medskip

\par\medskip\begin{center}
\includegraphics[width=\linewidth]{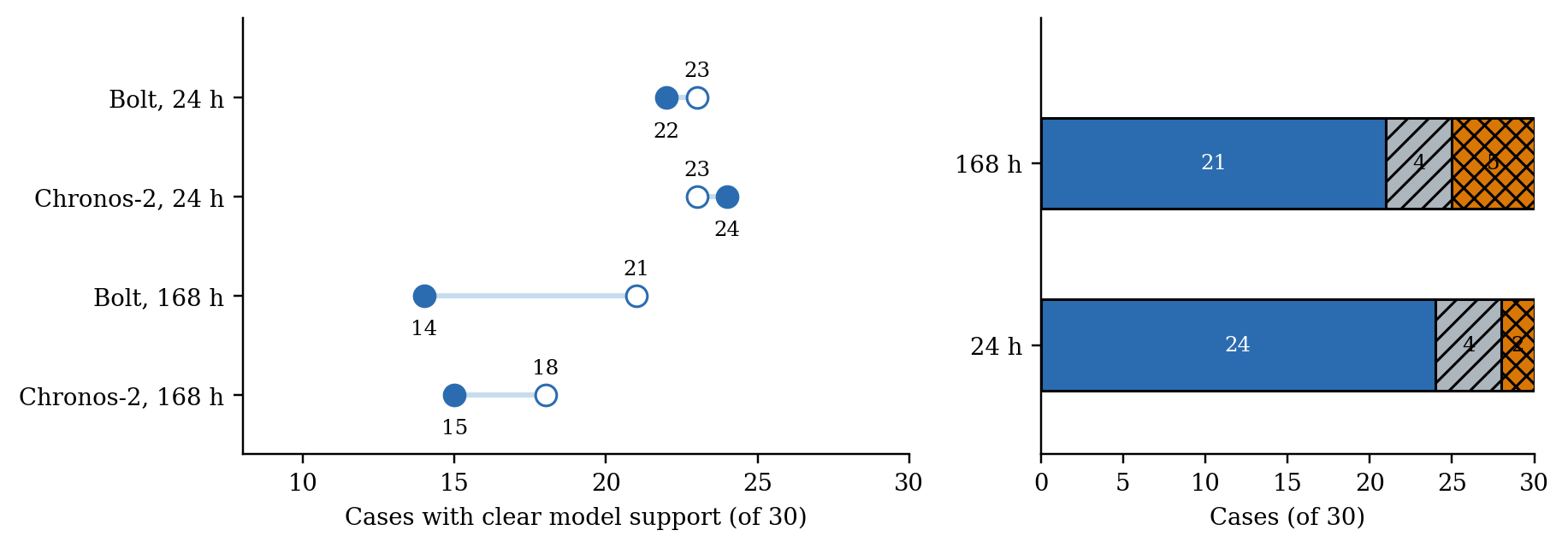}
\end{center}
\noindent\textbf{Figure 2:} Left: number of cases with clear paired support against weekly persistence (open circles) and against the validation-selected reference (filled circles). Right: test MAE of the selected reference compared with weekly persistence, lower (dark), equal (diagonal hatch) or higher (crosshatch). Each row covers 30 cases.
\par\medskip

\subsection{Chronos-2 versus Chronos-Bolt}

Chronos-2 has lower raw MAE than Bolt in 29 of 30 day-ahead cases and 24 of 30 week-ahead cases. Fewer of these differences are clear in the paired bootstrap. With both 168- and 336-hour blocks, Chronos-2 is clearly better in 15 day-ahead and 12 week-ahead cases. A week ahead, Bolt is clearly better in two cases, 14 are unclear or depend on the block length, and two have too little data. Chronos-2 therefore has a modest advantage that is not consistent across cases. Both models also come from the same family, so their agreement tells us little about foundation models built in other ways.

\par\medskip\begin{center}
\includegraphics[width=\linewidth]{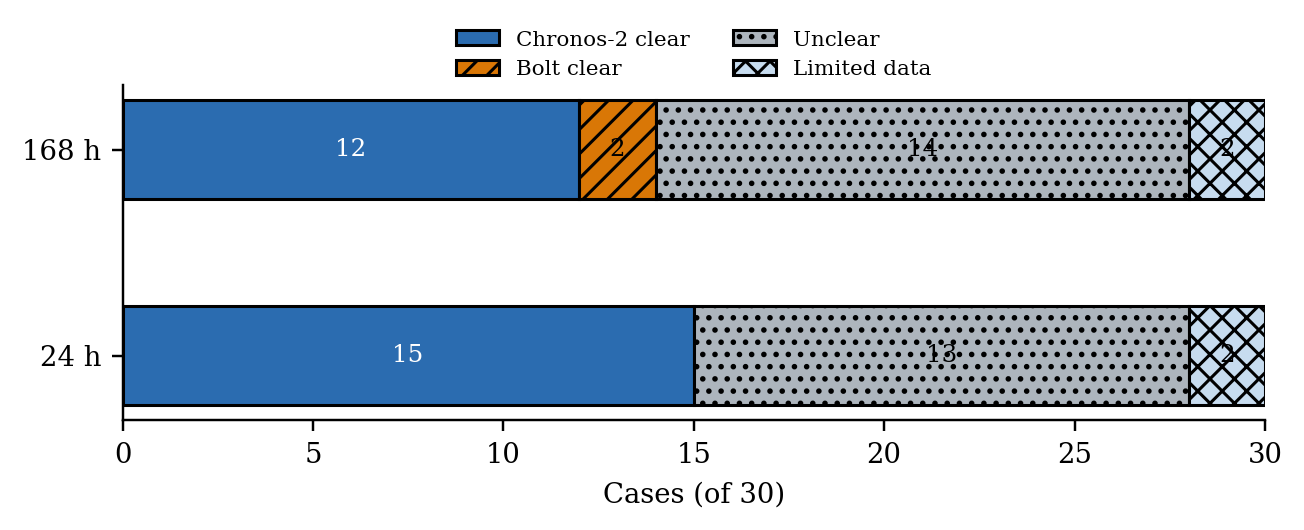}
\end{center}
\noindent\textbf{Figure 3:} Paired comparison of Chronos-2 and Bolt over the 30 cases at each horizon. A case is clear when both bootstrap block lengths favour the same model; cases without enough valid origins are shown separately.
\par\medskip

\subsection{A concrete Tetouan example}

Tetouan Zone 3 in Q4 illustrates the effect. Weekly persistence has an MAE of 921.0 there and the selected reference 607.4, while Bolt scores 543.8 and Chronos-2 scores 523.3. Against weekly persistence Bolt looks about 41.0\% better, and against the selected reference only 10.5\%. For Chronos-2 the gain falls from about 43.2\% to 13.8\%. The forecasts are identical in both comparisons; only the reference has changed. Q4, however, repeats the pilot period we had already inspected, so this case illustrates the point without confirming it. Zone 1 in the same quarter shows the opposite case. There the selected reference does worse than weekly persistence on test (865.8 against 814.0), which shows the risk of choosing a reference on validation data.

\subsection{Near-constant meter and suspected exposure}

We keep Portugal T3 because the meter was chosen before we looked at any test errors. An audit showed that 99.4\% of its 2014 hours exactly equal the value one week earlier in the published hourly series. Weekly persistence can therefore reach an MAE of zero or close to it, and a model's percentage gain becomes very large and negative, or undefined even when its absolute error is small. We report this series separately, keep it in the counts, and do not let its ratios stand for typical meters. The Portugal meters also come from a processed version of ElectricityLoadDiagrams20112014, which may overlap Chronos pretraining data. A chronological split on our side cannot exclude that, so for these series our results describe forecasting without fine-tuning and not a verified test on unseen data.

\subsection{Diagnostic comparison with trained models}

The validation-selected Ridge or LightGBM model has lower test MAE than the selected seasonal reference in 7 of 30 day-ahead cases and 4 of 30 week-ahead cases (Table 3). It has lower MAE than either Chronos model in 4 of 30 cases at each horizon, all four being Portugal T3.

\par\medskip
\begin{center}
\small
\setlength{\tabcolsep}{4pt}
\begin{tabular}{@{}lrrrr@{}}
\toprule
Horizon & Trained < selected seasonal & Trained < Bolt & Trained < Chronos-2 & Clear Chronos-2 over trained \\
\midrule
24 h & 7 / 30 & 4 / 30 & 4 / 30 & 24 / 30 \\
168 h & 4 / 30 & 4 / 30 & 4 / 30 & 21 / 30 \\
\bottomrule
\end{tabular}
\end{center}
\noindent\textbf{Table 3:} Case counts for the validation-selected Ridge or LightGBM model. Lower test MAE and clear paired support are different criteria. The four trained-model wins at either horizon are all Portugal T3.
\par\medskip

Outside that near-repeating meter, the selected trained model loses to the selected seasonal reference in 19 of 26 day-ahead cases and 22 of 26 week-ahead cases. In the paired bootstrap, each Chronos model has a clear win over the selected trained model in 24 day-ahead and 21 week-ahead cases, while the trained model is clearly favoured only in the four T3 cases at each horizon. These results show that this limited local training setup is not a consistently strong comparator. They do not establish that either Chronos model beats well-trained local forecasting methods in general.

\section{Discussion and Threats to Validity}

Changing the seasonal reference does not touch the Chronos forecasts, but it changes the size and certainty of the gains we can report. Against the selected rule, the median reported gain is lower for both models, and a week ahead there are fewer clear wins. At the same time, some individual comparisons become stronger, and the selected rule sometimes loses to weekly persistence on test. A reported gain should therefore always be read together with the baseline it was measured against.

The study has several limitations. First, the 60 cases come from only eight series in four collections, and the Tetouan zones and the Portugal meters are related to each other. Forecasts from hourly origins overlap, a complete test window holds only about five independent week-sized blocks, and the circular bootstrap assumes that losses behave roughly the same throughout a window. Second, Q4 repeats the pilot period, and the four-window design was set after the pilot had been inspected. Third, both models come from one family, and pretraining exposure has not been ruled out. Fourth, the local trained-model comparison uses only Ridge and LightGBM with limited settings and was designed after test inspection. Most of these fitted forecasts are weaker than the selected seasonal reference, so they cannot support a broad claim about Chronos against strong trained models. Fifth, the per-case intervals are not adjusted for multiple comparisons.

A stronger test would fix a new load period in advance and add a foundation model from a different family, with the same baseline rule, origins and uncertainty settings decided before any test forecast is looked at. That would show whether the pattern holds beyond these Chronos models and these past periods.


\begin{thebibliography}{99}

\bibitem{ref1} L. Simeone. \emph{Time Series Foundation Models for Energy Load Forecasting on Consumer Hardware: A Multi-Dimensional Zero-Shot Benchmark.} arXiv:2602.10848, 2026. \url{https://arxiv.org/abs/2602.10848}


\bibitem{ref2} M. Meyer, D. R. Zapata Gonzalez, S. B. Kaltenpoth, and O. Mueller. \emph{Benchmarking Time Series Foundation Models for Short-Term Household Electricity Load Forecasting.} IEEE Access 13, 218141-218153, 2025. \url{https://doi.org/10.1109/ACCESS.2025.3648056}


\bibitem{ref3} A. Vasulkar, S. Kernan Freire, A.-W. de Leeuw, and T.-B. Salcedo-Rahola. \emph{Machine Learning and Time-Series Foundation Models for Day-Ahead Electricity Forecasting in a University Building.} arXiv:2610.12638, 2026. \url{https://arxiv.org/abs/2610.12638}


\bibitem{ref4} M. Obermeier, M. Pruckner, F. Haselbeck, and A. Zeiselmair. \emph{FETS Benchmark: Foundation Models Enable Scalable and Generalizable Energy Time Series Forecasting.} Energy and AI, 100867, 2026. \url{https://doi.org/10.1016/j.egyai.2026.100867}


\bibitem{ref5} D. Yang. \emph{Making Reference Solar Forecasts with Climatology, Persistence, and Their Optimal Convex Combination.} Solar Energy 193, 981-985, 2019.


\bibitem{ref6} J. Lago, G. Marcjasz, B. De Schutter, and R. Weron. \emph{Forecasting Day-Ahead Electricity Prices: A Review of State-of-the-Art Algorithms, Best Practices and an Open-Access Benchmark.} Applied Energy 293, 116983, 2021. \url{https://doi.org/10.1016/j.apenergy.2021.116983}


\bibitem{ref7} UCI Machine Learning Repository. \emph{Power Consumption of Tetouan City.} \url{https://archive.ics.uci.edu/dataset/849/power+consumption+of+tetouan+city}


\bibitem{ref8} Monash Time Series Forecasting Repository. \emph{Electricity Hourly Dataset, version 3.} \url{https://zenodo.org/records/4656140}


\bibitem{ref9} Monash Time Series Forecasting Repository. \emph{Elecdemand Dataset, version 2.} \url{https://zenodo.org/records/4656069}


\bibitem{ref10} UCI Machine Learning Repository. \emph{Individual Household Electric Power Consumption.} \url{https://archive.ics.uci.edu/dataset/235/individual+household+electric+power+consumption}


\bibitem{ref11} A. F. Ansari, L. Stella, C. Turkmen, et al. \emph{Chronos: Learning the Language of Time Series.} Transactions on Machine Learning Research, 2024. \url{https://openreview.net/forum?id=gerNCVqqtR}. Chronos-Bolt weights: \url{https://huggingface.co/amazon/chronos-bolt-base}


\bibitem{ref12} A. F. Ansari, O. Shchur, J. Kueken, A. Auer, et al. \emph{Chronos-2: From Univariate to Universal Forecasting.} arXiv:2510.15821, 2025. \url{https://arxiv.org/abs/2510.15821}


\end{thebibliography}
\end{document}